\documentclass[letterpaper, 10 pt, conference]{ieeeconf}  

\IEEEoverridecommandlockouts                              

\usepackage{graphicx} 
\usepackage{amsmath} 
\usepackage{amssymb}  

\title{\LARGE \bf
MCTS Based Dispatch of Autonomous Vehicles under Operational Constraints for Continuous Transportation
}

\author{Milan Tomy$^{1}$, Konstantin M. Seiler$^{2}$, Andrew J. Hill$^{1}$
\thanks{$^{1}$ University of Sydney, NSW, Australia}
\thanks{$^{2}$ University of Technology Sydney, NSW, Australia}
}

\usepackage{amsfonts}
\DeclareMathOperator*{\argmax}{arg\,max}

\usepackage{algorithm}
\usepackage{algorithmic}
\usepackage{tikz}

\newcommand{\R}{\mathbb{R}}

\newcommand{\numtrucks}{\mathcal{H}}
\newcommand{\fasthc}{$\text{FAST}_{\text{HC}}$}
\newcommand{\fastcon}{$\text{FAST}_{\text{OC}}$}
\newcommand{\objt}{\mathbf{o}(t_{\text{e}})}
\newcommand{\objtc}{\bar{\mathbf{o}}(t_{\text{e}})}

\newcommand{\da}{d_\alpha}
\newcommand{\dela}{\delta_\alpha}

\begin{document}

\maketitle
\thispagestyle{empty}
\pagestyle{empty}

\begin{abstract}
Continuous transportation of material in the mining industry is achieved by the dispatch of autonomous haul-trucks with discrete haulage capacities. 
Recently, Monte Carlo Tree Search (MCTS) was successfully deployed in tackling challenges of long-run optimality, scalability and adaptability in haul-truck dispatch. 
Typically, operational constraints imposed on the mine site are satisfied by heuristic controllers or human operators independent of the dispatch planning.
This article incorporates operational constraint satisfaction into the dispatch planning by utilising the MCTS based dispatch planner Flow-Achieving Scheduling Tree (FAST).
Operational constraint violation and satisfaction are modelled as opportunity costs in the combinatorial optimisation problem of dispatch.
Explicit cost formulations are avoided by utilising MCTS generator models to derive opportunity costs.
Experimental studies with four types of operational constraints demonstrate the success of utilising opportunity costs for constraint satisfaction, and the effectiveness of integrating constraints into dispatch planning. 
%
\end{abstract}

%
\section{Introduction}
Automated dispatch strategies enable the efficient utilisation and growth of autonomous fleets in transportation and logistic applications such as public shuttle systems, automated warehouses, construction and mining.
A key aspect of automated dispatch is satisfaction of critical constraints imposed on the system to ensure safe and feasible operations.
%
Mine-sites are subject to a variety of operational constraints to ensure safety and satisfaction of personnel, prolong life-time of equipment and prevent disruptions to the flow and processing of material.
These constraints affect the dispatch decisions either by limiting the operational availability of vehicles in the fleet or by enforcing conditions on the requirement of material at locations.
For example, safe battery levels in battery electric trucks and safe tyre temperatures are imposed to avoid disruptions and hazards due to breakdown of the vehicle.
%
Maintenance routines and shift-change breaks restrict dispatch decisions within time windows to ensure safety and satisfaction of operators.

%
Typically, operational constraints are satisfied by conflict resolution after dispatch planning.
Heuristic controllers detached from dispatch planners, or interventions by human-operators correct any dispatch decision that would lead to a constraint violation \cite{hill2021mining}.
However, the deployment of such heuristic controllers require setting up of expensive infrastructure for enabling correction of dispatch decisions.
Thus, developing strategies that optimise haul-truck dispatch while accounting for operational constraint satisfaction could result in financial gains by preventing additional operational cost caused by correction.
It could also allow more cost-effective decisions regarding fleet size and equipment selection at the early mine planning stages \cite{leung2023automation}.

This article focuses on integrating operational constraint satisfaction into the haul-truck dispatch planning problem.
We consider four commonly encountered operational constraints varying in flexibility and specificity -- i) safe battery for haul-trucks; ii) safe tyre temperature for haul-trucks; iii) minimum volume of material at unloading stations and; iv) ratio of material volume at unloading stations.
%
Challenges arise from the extensive search space for finding feasible decision sequences, and complexity in mathematically formulating the effect of sequential decisions on operational constraint satisfaction.
The planner FAST \cite{Seiler2022} utilises MCTS to efficiently explore large search space and solve the haul-truck dispatch problem for continuous transportation.
The ability of FAST to optimise decisions over a future horizon, adapt to dynamic mine plans and provide real-time solutions for various fleet and mine sizes make FAST more advantageous than other popular dispatch algorithms in the mining domain.
%
Hence, we introduce an extension to FAST, denoted as \fastcon, to integrate operational constraint satisfaction into dispatch planning.
Our contributions include:
\begin{enumerate}
	\item Formulation of constraint violation and constraint satisfaction as opportunity costs.
	\item Incorporation of constraint satisfaction with opportunity costs into MCTS. Generator models are leveraged to estimate the opportunity costs, side-stepping the need for rigorous mathematical cost formulations.
	\item Comparison of \fastcon{} against FAST with independent heuristic controllers for constraint satisfaction, showcasing the improvements in terms of performance indicators in dispatch planning.
\end{enumerate}

\section{Related Work}
%
Haul-truck dispatch algorithms have evolved from the most popular greedy heuristics \cite{White2022} to mixed integer linear programming (MILP) solutions of instantaneous allocation problems \cite{Afrapoli2020} or evolutionary algorithms \cite{alexandre2019multi}, and eventually to reinforcement learning based methods such as MCTS (FAST \cite{Seiler2022}) and deep neural networks (DNN) \cite{Zhang2020, Joao2021}.
%
FAST and DNN methods address the previous shortcomings of long-term optimality and scalability.
However, FAST is more adaptable than DNN based methods as re-training is not required when production goals, mining locations and operational constraints change over time to accommodate evolving strategic mine plans.
%
Physical constraints affecting the mine and haul-trucks dynamics such as coordination and interaction at locations are accounted for in the algorithms mentioned above.
However, satisfaction of operational constraints are rarely considered. 

%
Though seldom studied in the mining domain, satisfaction of safety constraints in task planning has been considered in robotics.
%
A multi-robot system with a formal safety specification, similar to the nature of operational constraints in mining is studied in \cite{mansouri2019multi}.
The system is modelled with general stochastic petri network (GSPN) which allows specification of constraints as restrictions on the model. 
%
However, the GSPN formulation supports a specific type of constraint specification and produces offline solutions. 
Given a MILP formulation of a system, a variety of temporal constraints are incorporated into the dispatch planning of a heterogeneous team of robots in \cite{Leahy2022}. Off-the-shelf solvers are used to solve the MILP and generate offline policies.
Due to the difference in problem formulation and offline solutions, these constraint satisfaction strategies cannot be directly incorporated into online dispatch framework of FAST.

Within MCTS, constraint satisfaction can been incorporated by designing constraint-specific penalty functions for each violation \cite{Paxton2017}, or by modifying the selection policy to accommodate a constraint satisfaction heuristic \cite{aloor2023follow}. 
%
Both these methods require a knowledge of the objective function, to design penalty functions or estimate tuning parameters that bias MCTS exploration appropriately.
Instead, we design heuristic constraint-specific functions to generate an opportunity cost which avoids additional sensitive parameters that risk biasing the MCTS exploration.

\newcommand{\mine}{\mathcal{M}}
\newcommand{\tasks}{\mathcal{T}}
\newcommand{\roadnw}{\mathcal{R}}
\newcommand{\acts}{\mathcal{A}}
\newcommand{\conphy}{\mathcal{C}_p}
\newcommand{\conop}{\mathcal{C}_o}
\newcommand{\fconphy}{\mathbf{c}^p}
\newcommand{\fconop}{\mathbf{c}^o}

\section{System Description}
\label{sec:constraint_intro}
For the haul-truck dispatch problem, a simplified mine $\mine\!=\!(\numtrucks, \roadnw, \acts, \tasks, \conphy, \conop)$ is described by the set of haul-trucks $\numtrucks\!=\!\{v_1, v_2, \cdots\}$ available for operation, the road network graph $\roadnw\!=\!(L,E)$ determining the transportation path, the set of activities $\mathcal{A}$ defining all possible operational stages of a haul-truck, the set of haulage tasks $\tasks\!=\!\{T_1, T_2, \cdots\}$ defining the target rate at which material must be transported between locations, the set of physical constraint functions $\conphy\!=\!\{\fconphy_1, \fconphy_2, \cdots\}$, and the set of operational constraint functions $\conop\!=\!\{\fconop_1, \fconop_2, \cdots\}$.
%
Haul-trucks $v_i\!\in\!\numtrucks$ with a finite load capacity are dispatched to transport material between loading and unloading stations to achieve a continuous material flow rate set by a higher-level mine plan.
The set of materials $M\!=\!\{m_1, m_2, \cdots\}$ excavated at various loading stations at the mine include different grades of ore and waste.
While ore materials are transported to crushers and stockpiles for processing or blending to meet production requirements, waste material is transported to a dumpsite.
The traversable roads $e\!\in\!E$ between these locations $l\!\in\!L$ are determined by the directed weighted graph in road network $\mathcal{R}$.

The production goals of the mine are represented by continuous haulage tasks $T_j\!\!\in\!\!\mathcal{T}$.
Each task $T_j\!=\!(l_s, l_d, m_n, r)$ is defined by the transportation of material type $m_n\!\in\!M$ from loading station $l_s\!\in\!L$ to unloading station $l_d\!\in\!L$ at a target flow rate of $r\!\in\!\R$ tonnes per hour. 
Accomplishing haulage tasks requires repeated allocation of vehicles across the planning horizon.
On assignment of a task $T_j$ to haul-truck $v_i$, the following activities are executed in sequence by $v_i$ -- i) transiting empty to the excavator; ii) queuing at the excavator; iii) loading; iv) transiting loaded to the unloading station; v) queuing at the unloading station and; vi) unloading.
Additionally, haul-trucks are allocated actions $A_c\!=\!\{\textit{charge, park}\}$ for refuelling or recharging, maintenance and shift changes, to ensure operations around the clock.  
%
%
Then, the set of activity types $\bar{\acts}\!=\!\{\textit{transit}, \textit{load}, \textit{queue}, \textit{unload}, \textit{charge}, \textit{park}\}$ define the operational stages of a haul-truck at the mine-site.
Thus, an activity $\alpha\!=\!(l,\bar{\alpha})$ is uniquely determined by the activity type $\bar{\alpha}\!\in\!\bar{\mathcal{A}}$ executed at a location $l\!\in\!L$, resulting in the set of activities $\acts\!=\!\{\alpha | \hspace{0.1cm} \forall l \in L \text{ and } \forall \bar{\alpha} \in \bar{\acts}\}$.
Each dispatch decision or action results in the activity transition of a haul-truck.
The action duration of a haul-truck is equivalent to the duration $d_\alpha$ of the activity $\alpha\!\!\in\!\!\acts$ to which the truck transitioned.
Activity duration $d_\alpha$ is determined by $\bar{\alpha}$, $l$, the dispatch decisions and activity duration of other interacting vehicles, and domain specific interaction models.

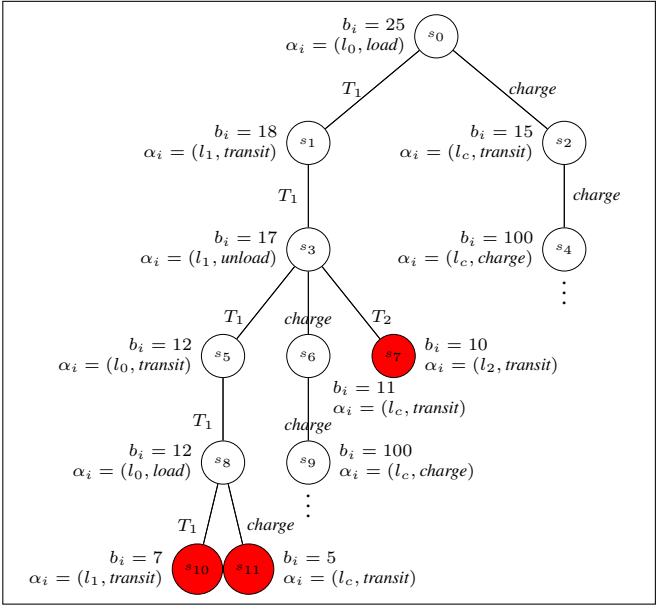
\begin{figure}[t]
	\centering
	\framebox{
		\resizebox{0.95\columnwidth}{!}{%
				
\begin{tikzpicture}[level/.style={sibling distance=36mm/#1, level distance=15mm}]
\tikzset{node style/.style={circle, draw, fill=white, text=black, align=center, font=\tiny}}

\node[node style, label={[align=right,font=\scriptsize]left:{$b_i=25$}\\$\alpha_i=(l_0, \textit{load})$}] (root){$s_0$}

child{node[node style, label={[align=right,font=\scriptsize]left:{$b_i=18$}\\$\alpha_i=(l_1, \textit{transit})$}] (c1) {$s_1$}
	child{node[node style, label={[align=right,font=\scriptsize]left:{$b_i=17$}\\$\alpha_i=(l_1, \textit{unload})$}] (c3) {$s_3$}
		child {node[node style, label={[align=right,font=\scriptsize]left:{$b_i=12$}\\$\alpha_i=(l_0, \textit{transit})$}] (c5) {$s_5$}
			child {node[node style, label={[align=right,font=\scriptsize]left:{$b_i=12$}\\$\alpha_i=(l_0, \textit{load})$}] (c8) {$s_8$}
				child {node[node style, label={[align=right,font=\scriptsize]left:{$b_i=7$}\\$\alpha_i=(l_1, \textit{transit})$}, fill=red] (c10) {$s_{10}$}
				}
				child {node[node style, label={[align=left,font=\scriptsize]right:{$b_i=5$}\\$\alpha_i=(l_c, \textit{transit})$}, fill=red] (c11) {$s_{11}$}
				}
			}
		}
		child {node[node style, label={[align=left,font=\scriptsize]below right:{$b_i=11$}\\$\alpha_i=(l_c, \textit{transit})$}] (c6)  {$s_6$}
			child {node[node style, label={[align=left,font=\scriptsize]right:{$b_i=100$}\\$\alpha_i=(l_c, \textit{charge})$}] (c9) {$s_9$}
				child[edge from parent/.style={draw=none}, level distance=5mm] {node (inv1){\vdots}}
			}
		}
		child {node[node style, label={[align=left,font=\scriptsize]right:{$b_i=10$}\\$\alpha_i=(l_2, \textit{transit})$}, fill=red] (c7) {$s_7$}
		}
	}
}
child{node[node style, label={[align=right,font=\scriptsize]left:{$b_i=15$}\\$\alpha_i=(l_c, \textit{transit})$}] (c2) {$s_2$}
	child {node[node style, label={[align=right,font=\scriptsize]left:{$b_i=100$}\\$\alpha_i=(l_c, \textit{charge})$}] (c4) {$s_4$}
		child[edge from parent/.style={draw=none}, level distance=5mm] {node (inv2){\vdots}}
	}
};

\draw (root) edge node[midway, left, font=\scriptsize, pos=0.5] {$T_1$} (c1);
\draw (root) edge node[midway, right, font=\scriptsize, pos=0.5] {\textit{charge}} (c2);
\draw (c1) edge node[midway, left, font=\scriptsize, pos=0.5] {$T_1$} (c3);
\draw (c2) edge node[midway, right, font=\scriptsize, pos=0.5] {\textit{charge}} (c4);
\draw (c3) edge node[midway, left, font=\scriptsize, pos=0.7] {$T_1$} (c5);
\draw (c3) edge node[midway, below, font=\scriptsize, pos=0.5] {\textit{charge}} (c6);
\draw (c3) edge node[midway, right, font=\scriptsize, pos=0.7] {$T_2$} (c7);
\draw (c5) edge node[midway, left, font=\scriptsize, pos=0.7] {$T_1$} (c8);
\draw (c8) edge node[midway, left, font=\scriptsize, pos=0.7] {$T_1$} (c10);
\draw (c8) edge node[midway, right, font=\scriptsize, pos=0.7] {\textit{charge}} (c11);
\draw (c6) edge node[midway, below, font=\scriptsize, pos=0.5] {\textit{charge}} (c9);

\end{tikzpicture}}
	}
	\caption{Violation of battery operational constraint $b_i > 10 \hspace{0.2cm} \forall v_i\in \numtrucks$ being caused by a sequence of decisions. Simplified state transitions of a single truck $v_i$ is depicted. Truck $v_i$ can choose among haulage tasks $T_1, T_2$ or \textit{charge} action. Red nodes indicate constraint violation.}
	\label{fig:opcost_eg}
	\vspace{-0.5cm}
\end{figure}

%
Physical constraints $\fconphy\!\in\!\conphy$ are imposed by $\roadnw, \numtrucks, \acts$ and the interaction model of the mine.
These are inviolable constraints which determine the dynamics of the system.
They constrain the volume of material moved by a vehicle, execution order of activities and activity duration.
Their statisfiability can be determined by monitoring the current state of the system.  
In contrast, operational constraints $\fconop\!\in\!\conop$ are non-markovian constraints which impact dispatch decisions without affecting the critical dynamics of the system.
Violation of operational constraints increase risk of safety hazards, equipment failure and consequent disruptions.
Unlike physical constraints, operational constraints cannot be satisfied by disallowing or enforcing actions in $\tasks \cup A_c$ based on a state, as the state from which a violation can occur cannot be easily identified.
A history or sequence of states is required to assess the constraint satisfiability.
From the simplified scenario in Figure \ref{fig:opcost_eg}, \textit{charge} action could be taken from $s_0$ or $s_3$ for constraint satisfaction, and states causing violation cannot be identified without complete forward simulation.
Thus, deciding an action based on a single state can lead to sub-optimal and conservative solutions.
This non-determinism is exacerbated due to haul-truck interactions as the decision of each truck and progress of constrained variables is dependent on the decisions of others.
Hence, haul-truck dispatch involves optimising allocation and scheduling of haulage tasks $\tasks$ and constraint actions $A_c$ to achieve production goals while satisfying the operational constraints of the mine-site.

\newcommand{\tend}{t_{\text{e}}}
\section{Problem Definition}
With the goal of developing a haul-truck dispatch planner considering operational constraints $\conop$, we address the problem of integrating operational constraint satisfaction into the MCTS based dispatch planner FAST.
%
The objective of haul-truck dispatch is to ensure continuous transportation of material to achieve production targets.
Let $\objt$ be a cumulative objective function $\mathbf{o} : \R \rightarrow \R$ that scores the progress of haulage tasks or goals until time $\tend$.
The cumulative objective $\objt$ is determined by the progress made by all actions or decisions $a_i^t$ of all haul-trucks $v_i\in\numtrucks$ until $\tend$.
Thus, haul-truck dispatch under operational constraints in a mine $\mine\!=\!(\numtrucks, \roadnw, \acts, \tasks, \conphy, \conop)$ with decision space $\mathbf{A}\!=\!\{\mathcal{T}\cup A_c\}$ is formulated as the following constrained sequential decision-making problem over a time horizon $H$:
\begin{equation}
\begin{aligned}
& \argmax_{a_i^t \in \mathbf{A}} && \mathbf{o}(H) \\
& \text{subject to} && \mathbf{c}^p(\mathbf{v}) \leq 0 &&& \forall \fconphy \in \conphy\\ 
& && \mathbf{c}^o(\mathbf{v}) \leq 0 &&& \forall \fconop \in \conop \\ 
\end{aligned}
\end{equation}
where $a_i^{t}\!\in\!\mathbf{A}$ represents the action or decision taken at time $t$ by haul-truck $v_i\!\in\!\numtrucks $, and  $\mathbf{v}$ is an $n$-dimensional vector representing system variables.
%
Operational constraint satisfaction depends on the sequence of decisions made by the multi-vehicle system as shown in the example in Figure \ref{fig:opcost_eg}.
The feasibility of an operational constraint can only be checked by monitoring the sequence of decisions chosen over a horizon.
Thus, the search space of feasible decisions grows exponentially with the action space of the vehicles, and the number of decisions made within a horizon which depends on the fleet size and length of planning horizon.
Hence, solution of the problem requires methods that allow efficient exploration of the extensive search space for objective maximisation within a limited execution time.

\section{FAST}
%
Flow-achieving Scheduling Tree (FAST) optimises haul-truck dispatch by reasoning over the sequence of system-wide dispatch decisions on a decision tree.
A decision tree node represents haulage task allocations of haul-trucks at time $t$.
FAST employs MCTS to efficiently explore the vast search space and optimise over the non-linear objective function in real-time.
MCTS is an anytime algorithm that grows the tree towards promising nodes by balancing exploration and exploitation based on a heuristic informed by random samples \cite{Kocsis2006, Silver2016}.

%
Each decision node encapsulates information for tracking progress of dispatch tasks, interactions among trucks and activity stations and decision node values.
%
The value of a decision node is estimated by forward simulation and random sampling with an efficient rollout policy. 
A high-performance state transition model capturing the interactions and asynchronous execution of vehicles allows speedy rollouts in MCTS.
Physical constraints are satisfied in FAST by disallowing actions causing violations and providing alternates that satisfy constraints in the state transition models. 
Each haul-truck $v_i\!\in\!\mathcal{H}$ is deterministically chosen by selecting the vehicle that finishes an action execution earliest.
Thus, the decision nodes in the tree represent decisions at sequential time points ordered such that earlier haul-truck decisions and activities are closer to the root.
For balancing exploration and exploitation, FAST utilises a self-tuned variant of UCB algorithm \cite{Kocsis2006} for the selection of action or haulage task  $a_i^t\!\in\!\mathcal{T}$ for haul-truck $v_i$ at time $t$ based on node values.

For estimating decision node values, FAST introduced a cumulative objective function $\objt$ that measures the overall deviation from targets defined in haulage tasks $T\in \mathcal{T}$, by tracking the cumulative volume of material moved by the system until time $\tend$.
A goal function $ \mathbf{g} : \mathcal{T} \times \R \rightarrow \R$ maps a task $T\! \in\! \mathcal{T}$ to the cumulative target volume of material $m_n\!\in\!M$ to be moved from loading station $l_s\!\in\!L$ to unloading station $l_d\!\in\!L$ by time $\tend$.
Similarly, a flow function $\mathbf{f} : L \times L \times \R \rightarrow \R$, defines the cumulative volume of material moved between a pair of locations $l_s$ and $l_d$ with decisions taken from the action space $\mathbf{A}\!=\!\mathcal{T}$ until time $\tend$.
Thus, the simplified cumulative objective function is:
\begin{equation}
\label{eq:obj_t}
\objt = \sum_{T\in\mathcal{T}}{\mathbf{e} \left( \mathbf{f}(l_s,l_d, \tend) - \mathbf{g}(T, \tend) \right)} 
\end{equation}
where $\mathbf{e}$ is an application specific error function designed to penalise deviation from target material flow rates. 
Hence, $\objt$ depends on haulage task completion duration, which is determined by the activity durations of the haul-trucks performing the task.
Exponential discounting with a discount factor $\zeta\!\in\!(0,1)$ is incorporated to bound the objective $\mathbf{o}(t)$ over an infinite time horizon, and prioritise immediate rewards over those obtained later into the horizon.
%
The discounted objective is calculated across uniformly discretised timesteps $\Delta t$ and is given by $\mathbf{o}^{\zeta}(n\Delta t) = \mathbf{o}(0) + \sum_{i=1}^{n}{\zeta^{i}\left(\mathbf{o}(i\Delta t)-\mathbf{o}((i-1)\Delta t)\right)}$.

%
FAST executes the decision from the root that maximises the objective and re-runs MCTS for obtaining decisions at later time points.
Thus, by quickly re-planing over a fixed horizon with receding horizon control, FAST approximates long-term optimality, and accounts for uncertainty and disruptions without explicit stochastic optimisation.

\newcommand{\dv}{\mathbf{d}^\text{v}}
\newcommand{\ds}{\mathbf{d}^\text{s}}
\newcommand{\fv}{f_c^\text{v}}
\newcommand{\fs}{f_c^\text{s}}

\section{Operational Constraint Satisfaction}
\label{sec:opp_costs}
%
The intractable search space for decisions feasible under operational constraints poses the challenge in dispatch planning with constraint satisfaction.
We tackle this challenge by leveraging MCTS and opportunity costs estimated from MCTS generator models for efficient search.
\subsection{Opportunity cost}
Cost functions penalise the objective on constraint violation and direct the search towards higher-valued decisions which also satisfy constraints.
In our problem, designing costs for violations is challenging due to the interdependence of haul-truck decisions and incomplete knowledge on the range of the objective function $\objt$.
For example, estimating the cost of battery constraint violation in Figure~\ref{fig:opcost_eg} is difficult as any loss incurred could be accounted for by other vehicles in the system.
Assigning costs for such violations without knowing the range of objective values can reduce MCTS rewards and bias MCTS exploration away from nodes propagating low reward values, resulting in sub-optimal solutions.
Hence, opportunity costs provide a suitable framework for comparing decision values without explicitly formulating cost functions or knowing the complete reward distribution.

An opportunity cost of a decision is the value forgone \cite{buchanan1991opportunity}. 
Here, opportunity costs are the rewards forgone by executing actions that do not contribute towards task progression instead of actions that only contribute to task progression.
%
Actions that result in constraint violation incur opportunity costs by increasing the chance of disrupting task progression.
%
Actions $a_i^t\!\in\!A_c\!=\!\{\textit{charge, park}\}$ also incur opportunity costs as they delay task progression while aiding constraint satisfaction.
Thus, a cumulative opportunity cost can be defined as $\mathcal{OC}(\tend)\! =\! \objtc - \objt$, where $\objtc$ is the return on executing actions $a_i^t\!\in\!\{A_c\cup\tasks\}$ causing possible constraint violations and delays to task progression until $\tend$, and $\objt$ denotes the return on taking actions $a_i^t\!\in\!\tasks$ that only progress task execution without any constraint violations until $\tend$.
We avoid explicit formulation and tuning of cost functions by directly estimating the new objective $\objtc=\objt+\mathcal{OC}(\tend)$ with heuristic MCTS generator models.

\subsection{Generating opportunity costs}
\label{sec:opcost_gen}
For estimating the cumulative objective augmented with opportunity cost, $\objtc$, a suitable variable that creates a loss in $\objt$ on constraint satisfaction and violation must be identified.
In our context, $\objt$ is dependent on the sum of all activity durations $\da^{i,t}$, resulting from actions $a_i^t\!\in\!\mathbf{A}$ taken by vehicles $v_i\!\in\!\numtrucks$ at time points $t$, until $\tend$.
Thus, $\objt$ in Equation \ref{eq:obj_t} can also be represented as $\objt\!=\! f\left(\sum_{v_i\in\numtrucks}\sum_{t<\tend}\da^{i,t}.\mathbf{1}_{a_i^t\in\tasks}\right)$.
%
Activity duration $\da^{i,t}$ is determined by a function $g : \tasks\cup A_c\times \acts \times \R \rightarrow \R$ which maps an action $a_i^t$, the resulting activity $\alpha\!\in\!\mathcal{A}$, and nominal activity duration $D_\alpha\!\in\!\R$ to $\da^{i,t}\!\in\!\R$, based on domain specific models of the activity type $\bar{\alpha}$ and interaction among vehicles.
Here, $D_\alpha$ is the nominal duration of the activity without interactions obtained from the data on mine operations. 
Hence, the activity duration is given by $\da^{i,t} = g(a_i^t, \alpha, D_\alpha)$.

%
When an action $a_i^t\!\in\!\mathcal{T}\cup A_c$ leads to constraint violation, its activity duration $\da^{i,t}$ can be manipulated by adding a delay $\dela^{i,t}$ to impose a loss.
Thus, the cumulative opportunity cost $\mathcal{OC}(t)$ is generated by the cumulative delay in achieving targets imposed by actions that violate constraints and procrastinate task completion.
%
The opportunity cost inclusive cumulative objective $\objtc$ at the end of planning horizon $H$ is estimated as follows:
\begin{equation}
\label{eq:opcost_delays}
\begin{aligned}
\bar{\mathbf{o}}(H) &= \mathbf{o}(H) + \mathcal{OC}(H)\\
&= f\left(\sum_{v_i\in\numtrucks}\sum_{t<H}\da^{i,t}.\mathbf{1}_{a_i^t\in\tasks}\right) + \mathcal{OC}(H)\\
&= f\left(\sum_{v_i\in\numtrucks}\sum_{t<H}(\da^{i,t}+\dela^{i,t}).\mathbf{1}_{a_i^t\in\tasks\cup A_c}  \right)
\end{aligned}
\end{equation}
The delays incorporated affect function $\mathbf{f}$, the cumulative material moved in Equation \ref{eq:obj_t}, allowing us to estimate $\objtc$ with the same formulation.
%
No opportunity costs are incurred when an action $a_i^t$ does not result in constraint violation as $\dela^{i,t}=0$, and when no actions $a_i^t \!\in\!A_c$ are taken. 
Including opportunity costs for constraint satisfaction allows the planner to reason among different actions that satisfy the same constraint, and trade-off between competing constraint satisfaction and objective maximisation.
Delay $\dela^{i,t}$ incurred on violating a constraint controls the rate at which cost accumulates in $\objtc$.
Hence, we design delay generator functions to estimate $\dela^{i,t}$ and approximate the augmented cumulative objective $\objtc$.

\subsection{Designing violation delay}
\label{sec:opcon_details}
Operational constraints considered in this system can be classified based on i) flexibility -- hard or soft constraints; ii) specificity -- the constraint is satisfied individually by each vehicle in the fleet or collectively by only a subset of the fleet.
The violation delay $\dela^{i,t}$ incurred is dependent on the type of operational constraint considered. 
We define an infinite delay for the violation of hard constraints $\mathcal{C}_o^{h}$ since it continuously accumulates error over the horizon, as shown in Equation \ref{eq:obj_t}, leading to a diminishing the value of $\objtc$.
In contrast, $\dela^{i,t}$ is defined by a function $f_c^v:\R^n \rightarrow \R$ for soft constraints $\mathcal{C}_o^{s}$, such that the delay is proportional to the error in constraint satisfaction.
Thus, constraint violation delay resulting from action $a_i^t$ is:
\begin{equation}
\label{eq:violationdelay}
\begin{aligned}
& \dela^{i,t} = \fv(\mathbf{v}^c) && \fconop \in \mathcal{C}_o^{s} \subseteq \conop\\
& \dela^{i,t} =   \infty  && \fconop \in \mathcal{C}_o^{h}\subseteq \conop\\
\end{aligned}
\end{equation}
where $\fv$ is a constraint-specific function that generates delay based on the constrained variable $\mathbf{v}^c$.
For vehicle-specific constraints, the violation delay $\dela^{i,t}$ is incurred on a single vehicle $v_i$ that violated the constraint.
However, for collective constraints, the $\dela^{i,t}$ is incurred by all trucks executing an activity $\alpha\!\in\!\acts$ specific to the constraint violation.
%
Violation delay $\dela^{i,t}$ represents delays incurred when constraint violation results in the worst-case or undesirable behaviour and need not accurately estimate the dynamics of the system.
Next, we design delay function $f_c^\text{v}$ for four different types of operational constraints typical for a mine-site. 

\subsubsection{Battery constraints} 
Battery life restricts the available operational time of battery electric haul-trucks. 
%
Let $b_i\!\in\![0,100]$ denote the battery of haul-truck $v_i$.
Then, the constraint function is $b_i>B_{\min} \hspace{0.1cm} \forall v_i\!\in\!\numtrucks$, with constant $B_{\min}$. 
These are hard, vehicle-specific constraints imposed to prevent disruptions due to haul-truck breakdown.
We assume a recharge system with limited charging stations at a charging bay, and linear charging and discharging models.
Updated battery $b_i'$ is given by $b_i' = b_i + k_{\alpha}\da^{i,t}$, where $k_\alpha$ is the charging rate or activity dependant discharging rate, and $\da^{i,t}$ is the activity duration for any action $a_i^t$. 
The haul-truck is assumed to charge till maximum capacity when $a_i^t\in A_c$ results in a charging activity, resulting in a delay imposed by charging duration $d_{\alpha}^{i,t} = \frac{100-b_i}{k_\alpha}$ obtained from the charging model.
%
Regardless of the battery model, infinite violation delay $\dela^{i,t}$ for each truck violating the hard constraint emulates the effect of a truck being inoperable when battery constraints are violated. 
\subsubsection{Tyre temperature constraints}
Short dispatch tasks decrease the overall tyre temperature, while long tasks result in an overall increase.
Thus, allocation of tasks to haul-trucks should maintain safe tyre temperature to avoid life-threatening hazards from tyre bursting.  
These can be hard or soft constraints that must be satisfied by each vehicle.
Let $y_i\!\in\![35,95]$ denote the tyre temperature of haul-truck $v_i$.
The soft constraint function is $y_i<Y_{\text{Th}} \hspace{0.1cm} \forall v_i\!\in\!\numtrucks$, where $Y_{\text{Th}}$ is the constant soft threshold to be maintained. 
The heating of tyres is modelled with linear function $y_i' = y_i + k_{h_{\alpha}}\da^{i,t}$ and cooling of tyres is given by the exponential function $y_i' = 35^\circ \text{C} + y_ie^{-k_{c}\da^{i,t}}$, where $k_{h}$ is the constant heating rate, $k_c$ is the constant cooling rate of the tyre and $35^\circ \text{C}$ is the ambient temperature.
Tyres heat for any action leading to an activity type $\bar\alpha\! =\!\textit{transit}$ and cool otherwise.
An opportunity cost for satisfying constraints is only imposed when $a_i^t= \textit{park}$ and is determined by the tyre cooling model $\da^{i,t}\!=\! \frac{1}{k_c}\ln\frac{y_{\text{final} - 35}}{y_i}$.
%
Violation delay $\dela$ of the soft constraint $y_i < y_\text{Th}$ resulting from an action $a_i^t$ for a truck $v_i$ is generated by the function:
\begin{equation*}
\dela^{i,t} = f_c^v(y_i) =
\begin{cases} 
K\da^{i,t}, &\text{if } \bar{\alpha} = \textit{transit}\\
0, & \text{otherwise}
\end{cases}
\end{equation*}
where $K$ is a scaling factor that can be constant or proportional to the violation error.
In our experiments $K=1$, which models the truck slowing down when transiting with hot tyres, imposing a delay on task progress.
No delays are imposed while the truck is cooling down.
\subsubsection{Capacity constraints}
\label{sec:crusher_dynamics}
Minimum capacity at locations is modelled as a hard, collective constraint-specific to an unloading activity.
These operational constraints define the minimum volume of material $V_{\min}$ required at unloading stations such as a crusher.
When a crusher is operated with volume of material $V(t) < V_{\min}$, the components of the crusher are prone to damage, increasing the likelihood of crusher failure.
We assume the crusher processes material at a constant rate $p$ tonnes per hour. 
The volume of material in the crusher $V(t)\!\in\!\R^+$ increases when an action $a_i^t$ results in the activity $\alpha = (\textit{crusher, unloading})$ and continuously decreases at a rate $p$ till the next unload activity.
%
As minimum capacity constraints must be collectively satisfied, no particular $v_i$ causes the constraint violation.
%
Thus, when the constraint is violated, i.e., $V(t) \leq V_{\min}$, the violation delay $\dela^{i,t}=\infty$ is augmented for any truck $v_i$ performing the specific activity $\alpha=(\textit{crusher, unloading})$ at time $t$. 
Infinite violation delay on these trucks emulates the effect of the crusher being inoperable and hindering the progress of all tasks involving the crusher. 
None of the constraint-satisfying actions $a_i^t \in \mathcal{T}$ incur opportunity costs. 
%
\subsubsection{Ratio constraints}
These are soft and collective constraints defining the required ratio among the volume of materials at a location.
Ratio constraints are imposed at the crusher for blending different grades of ore or for limiting the volume of sticky or challenging material to avoid clogging the crusher and risk of failure.
Hence, ratio constraints determine the sequencing of dispatch decisions at locations.
Assuming the crusher dynamics as in Section \ref{sec:crusher_dynamics}, let $V_{m_j}(t)$ denote the volume of material type $m_j$ in the crusher bin, $V(t)$ be the total volume of material in the crusher, $V_{\max}$ tonnes be the maximum bin capacity of the crusher and $p$ be the processing rate.
As the errors in required ratios $e_{m_j}(t)$ increase, the processing rate of the crusher is reduced exponentially to $pe^{-\max_j e_{m_j}(t)}$, to model the gradual blockage and ultimate breakdown of crusher. 
The error in required ratio $r_{m_j}^*$ for volume of material $m_j$ is given by $e_{m_j}=|\frac{r_{m_j}(t)-r_{m_j}^*}{r_{m_j}^*}|$, where $r_{m_j}(t) = \frac{V_{m_j}(t)}{V_{\max}}$.
As the processing rate slows down, the total volume in the crusher bin $V(t)$ approaches the maximum $V_{\max}$ delaying unloading activity and progress of any task involving the crusher.
Thus, the violation delay is proportional to error in constraint satisfaction and is augmented to all trucks interacting at the location.
The action $a_i^t$ by vehicle $v_i$ at time $t$ results in a delay $\dela^{i,t}$ only when the $a_i^t$ causes the activity $\alpha=(\textit{crusher, queue})$ and is given by:
\begin{equation*}
\dela^{i,t} = 
\begin{cases}
0,  \text{ if }C_i + V(t_{q}) \leq V_{\max}\\
\left(C_i + V(t_{q}) - V_{\max}\right)\left(\frac{1}{p}-\frac{1}{pe^{-\max_j e_{m_j}(t_q)}}\right)
\end{cases}
\end{equation*}
which is the additional duration for the crusher to process volume equivalent to the capacity $C_i$ of truck $v_i$ before unloading.
The delay and error is dependent on time $t_q$ denoting the time at which $v_i$ reaches the front of the queue.

\subsection{FAST with operational constraints -- \fastcon}
\label{sec:fast_with_constraints}
%
The activity durations $\da^{i,t}$ and constraint violation delays $\dela^{i,t}$ are utilised to formulate the opportunity cost inclusive objective $\objtc$ in Equation \ref{eq:opcost_delays} while maintaining the same formulation of objective $\objt$ introduced in FAST (Equation \ref{eq:obj_t}). 
Thus, we maintain the entire framework of FAST including the state and action space, state transition and interaction models, MCTS characterisations and online receding horizon execution.
The state space is augmented with constrained variables required for monitoring constraint violation, and action space includes constraint specific actions $A_c$ in addition to tasks $T \in \mathcal{T}$.
The state transition model includes dynamics and delays for calculating the cost inclusive cumulative objective as described in Section \ref{sec:opcon_details}.

%
The rollout policy in MCTS may have to select among task specific actions $a_i^t \in \mathcal{T}$ and constraint specific actions $a_i^t \in A_c$.
Randomly sampling from the action space $\mathcal{T}\cup A_c$ with a uniform probability distribution results in noisy rollouts and un-informed exploration in MCTS.
To ensure unbiased exploration and exploitation, a rollout policy should make random choices with realistic outcomes \cite{gelly2007combining}.
Uniform sampling selects actions $a_i^t \in A_c$ too often when they are not required and accumulates opportunity costs.
These low rewards bias the exploration resulting in sub-optimal decisions.
Thus, to model realistic outcomes, the probability of selecting a constraint-specific action $p(a_i^t \in A_c) = h(\mathbf{v}^c)$ is determined by the closeness of the state variable to constraint violation.
For example, the probability of charging is given by $p(\textit{charge}) = \frac{100-b_i}{100-B_{\min}}$ for battery constraints $b_i > B_{\min} \hspace{0.2cm} \forall v_i\in\numtrucks$.
Then, the probability of task execution is $p_\text{execute} = 1-p(\textit{charge})$, and the probability of selecting a task $T_j\in\tasks$ proportional to the required flow rate as in FAST.
For constraints with no new actions, the rollout heuristic in FAST is sufficient as there is no additional opportunity cost for constraint satisfaction.

\newcommand{\horizon}{$H$}
\newcommand{\halftime}{$H_{0.5}$}
\section{Empirical Analysis}
%
Each operational constraint is studied individually without the influence of the other constraints on the system to experimentally evaluate the ability of \fastcon{} in satisfying operational constraints while optimising the objective.
\subsection{Experimental Setup}
All experiments were run in simulation and built on a miniature toy mine $\mathcal{M}$.
\fastcon{} was implemented on python and simulated with a state-based simulator on a desktop computer with Intel Core i7-6700K CPU and 16GB of RAM.
\subsubsection{Environment}
\begin{figure}[]
	\centering
	\framebox{\includegraphics[width=0.96\columnwidth]{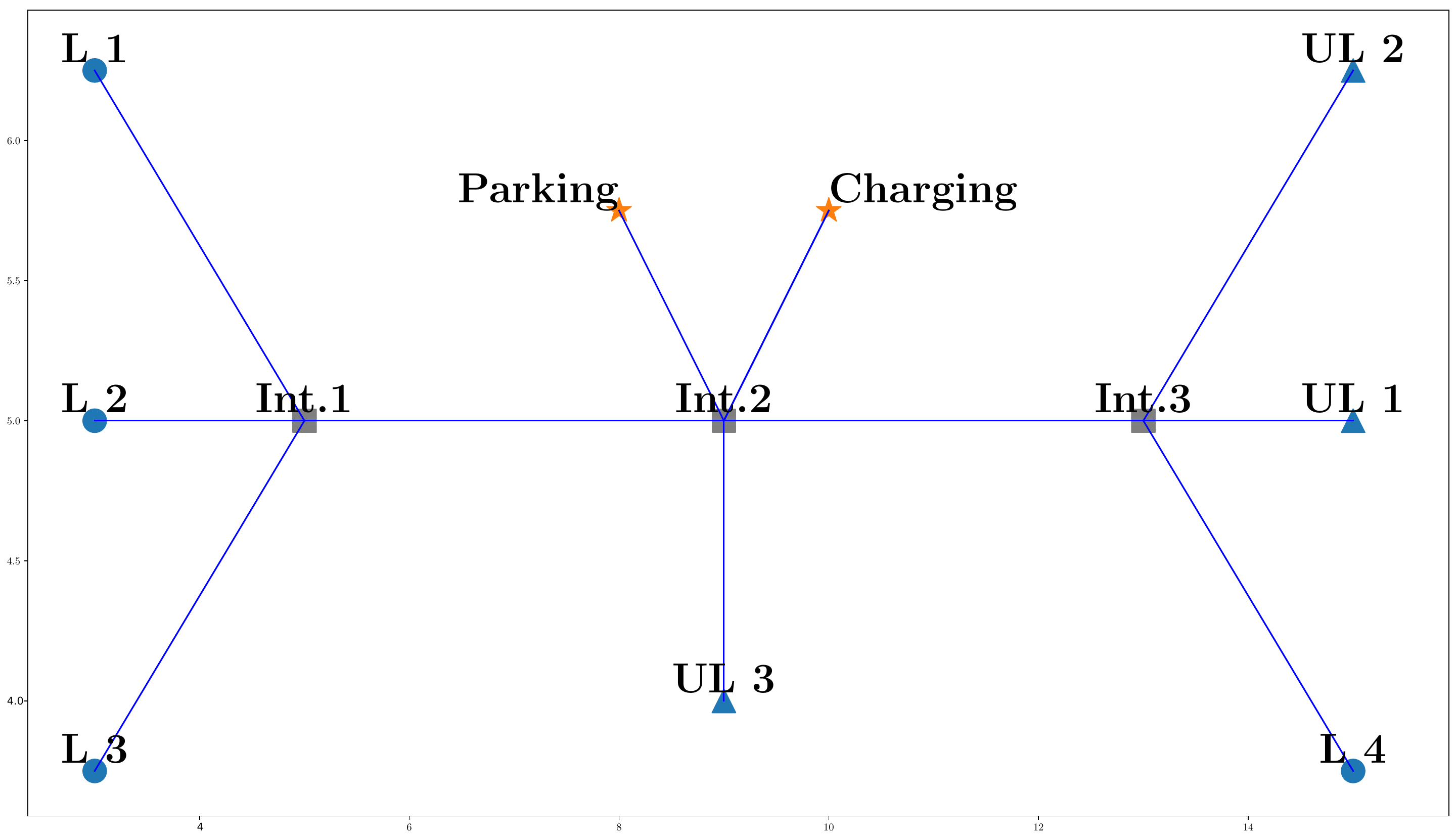}}
	\caption{Road network $\mathcal{R}$ where $\text{L}_i$ are loading and $\text{UL}_i$ unloading stations. The scale of the graph edges reflects the transit duration on each edge.}
	\label{minemap}
	\vspace{-0.5cm}
\end{figure}

The road network $\mathcal{R}$ used for the experiment is shown in Figure \ref{minemap}.  
A set of four dispatch tasks $\mathcal{T}$ has been described for $\mathcal{M}$: i) $T_0 = (\text{L}_1, \text{UL}_1, \text{M}_1, 100)$, which signifies the task of continuously transporting $\text{M}_1$ from $\text{L}_1$ to $\text{UL}_1$ at a target rate of 100 tonnes per hour; ii) $T_1 = (\text{L}_2, \text{UL}_2, \text{M}_2, 100)$; iii) $T_2 = (\text{L}_3, \text{UL}_3, \text{waste}, 100)$; iv) $T_3 = (\text{L}_4, \text{UL}_1, \text{M}_3, 200)$.
Note that $T_3$ is significantly shorter than other three tasks. 
We assume $\mathcal{H}$ to be a set of 5 haul-trucks with 100 tonnes capacity each. 
Without any operational constraints $\mathcal{C}_o$, $\mathcal{M}=(\mathcal{H,R,A,T,C}_p)$ represents a self-sufficient system capable of achieving targets. 

\subsubsection{Tuning parameters}
Optimal planning horizon $H$ and discount rate $\zeta$ for maximising objective score and satisfying constraints were obtained by analysing performance indicators for varying $H$ and constant $\zeta$, and vice-versa. 
Longest duration to observe reward $d_{r}$ is utilised to express $H$ and $\zeta$. 
The duration $d_{r}$ varies depending on the constraint imposed.
Planning horizon is expressed as $H=f_{hz}d_r$, where $f_{hz}$ is the horizon factor.
Halftime $H_{0.5}$ denotes the time to halve the objective value with $\zeta$. 
The relation between $H_{0.5}$ and $\zeta$ is given by $\frac{1}{2} = \zeta^{\frac{H_{0.5}}{\Delta t}}$, where $H_{0.5}$ can be expressed in terms of half-time factor $f_{hf}$ as $H_{0.5}= f_{hf}d_r$.
\subsubsection{Performance indicators}
The online plans generated are analysed based on the following major performance metrics of the mine-site:
i) Score of plan -- Score describes the deviation of total material moved from the targets. It is defined by the error function used in Equation \ref{eq:obj_t}. A positive score indicates that the target flow rates for tasks are maintained or exceeded; 
%
iii) Operational time -- The total fleet duration for \textit{loading}, \textit{unloading} and \textit{transiting}; 
iv) Queuing time; and other constraint-specific durations indicating the total fleet duration for constraint satisfaction or violation.
%
\subsubsection{Baseline planners \fasthc}
Typically, each operational constraint is handled at the mine-site with a heuristic controller detached from the dispatch planning process. 
Thus, a fully tuned FAST with no constraints and independent heuristic controls for constraint satisfaction are developed as a baseline.
The heuristics utilised are described briefly:
\paragraph{Battery \& tyre temperature heuristic}
For all possible tasks from the current state, the heuristic deterministically looks ahead to evaluate if executing a partial task and taking a constraint-satisfying action in $A_c$ will result in a hard constraint violation. 
If the look-ahead check indicates a constraint violation for any of the tasks, then the conservative heuristic decides on a constraint-satisfying action.
%
\paragraph{Capacity heuristic}
A minimum number of trucks must always execute the task corresponding to the location where the minimum capacity constraint has been imposed. 
\paragraph{Ratio heuristic}
The volume ratio of the material with lower proportion is maintained lesser than or equal to specified constraint ratio by assigning arriving trucks to a buffer queue when their contribution violates the ratio.
\subsection{Results \& Discussion}
Results presented are the averages across 15 simulations, with same initial conditions.
Online plans were generated for one day and analysed to obtain key performance indicators for each experiment.
For each online decision, MCTS was run with 10,000 iterations, and $\Delta t$ discretisation for reward calculation was set to 600s.
\subsubsection{Battery constraints}
\begin{figure*}[t]
	\centering
	\framebox{\includegraphics[width=1.98\columnwidth]{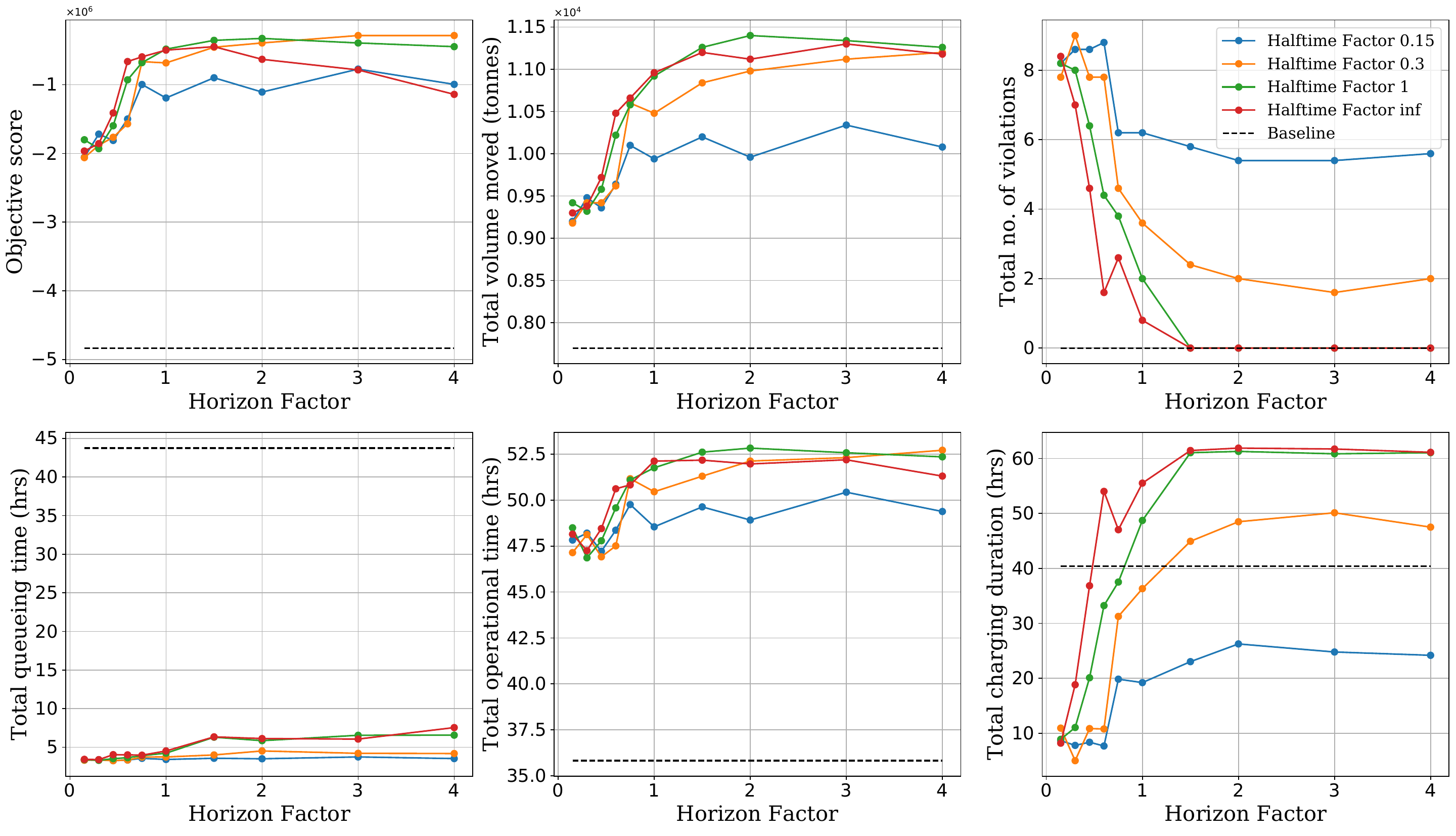}}
	\caption{Comparing \fastcon{} with battery constraint against \fasthc{} for varying $f_{hz}$ and fixed $f_{hf}$. Horizon $H\!=\!7f_{hz}$ hrs, Halftime $H_{0.5}\!=\!7f_{hf}$ hrs}
	\label{kpibattery}
	\vspace{-0.5cm}
\end{figure*}
$\forall v_i \in \numtrucks : b_i > B_{\min}$

Tuning results of \fastcon, for varying the horizon is presented in Figure \ref{kpibattery}.
Performance in terms of objective score, material moved and constraints satisfied is maximised as \horizon{} increased to 10.5 hours and \halftime{} increased to 7 hours ie; discount factor of 0.98.
This is because MCTS makes decisions by trading-off more accurate representations of opportunity costs. 
Our battery model restricts the operational availability of a vehicle to about 3-4 hours with a fully charged battery, and requires about 4 hours to fully charge.
Thus, MCTS rollouts can only estimate the opportunity cost of satisfying the constraint by looking beyond 8 hours. 
For small horizons, low charging durations indicate that few \textit{charge} decisions were made to avoid constraint violations.

A decrease in objective is noticed for \horizon{} greater than 10.5 hours and discount factors of 1 or \halftime$=\infty$. 
When $\zeta$ is 1, rewards accomplished early and later into the horizon are equally weighted. 
Thus, the controller is unable to prioritise avoiding delays and accomplishing higher rewards sooner. 
This leads to an increase in queuing time.
It also tends to prioritise shorter tasks over longer ones as they prolong the operational time by using lesser battery life, enabling greater quantity of tasks to be done within a horizon.
Longer horizons with no discounting tends to prefer quantity over quality, resulting in a significant decrease in objective score while maintaining the quantity of material moved.

Results show that best tuned \fastcon{} improves the cumulative material moved and operational time of \fasthc{} by about 45\%, while satisfying the hard constraint.
\fastcon{} accomplishes more work by charging opportunistically and queuing lesser. 
The queuing time of \fastcon{} is 84\% lower than that of \fasthc, as it coordinates with vehicles in the system and accounts for contention of resources. 
Though all vehicles started with similar battery levels, \fastcon{} scheduled charging such that queuing at the charging bay was minimised.
Additional rules such as ensuring staggered battery values is required to avoid such queuing with \fasthc.
\subsubsection{Tyre temperature soft constraints}
$\forall v_i \in \numtrucks : y_i < Y_\text{Th}$ 

\begin{figure}[ht]
	\centering
	\framebox{\includegraphics[width=0.96\columnwidth]{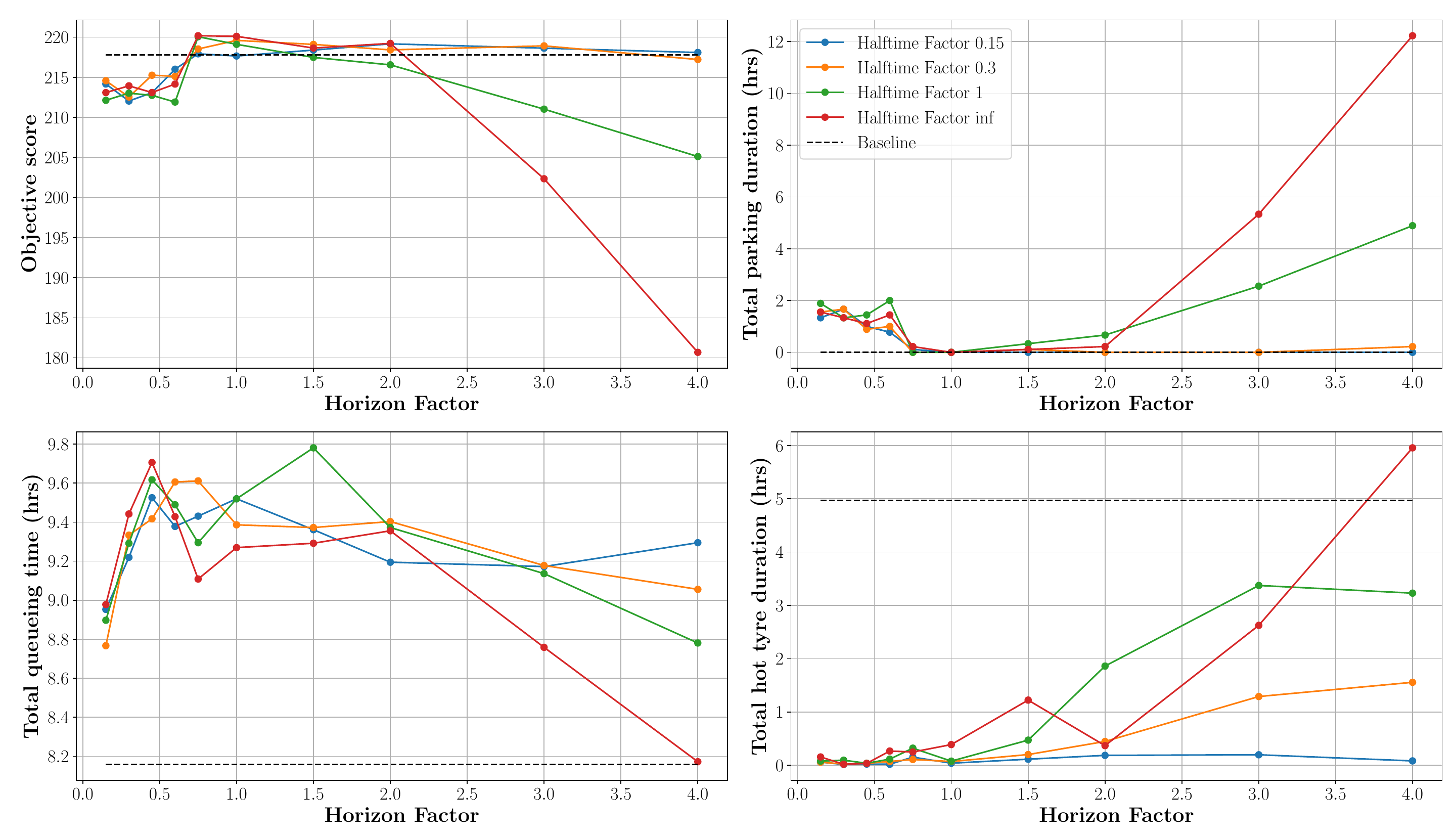}}
	\caption{Comparing \fastcon{} with tyre temperature constraint against \fasthc{} for varying $f_{hz}$ and fixed $f_{hf}$. $H=4f_{hz}$ hrs,  $H_{0.5}=4f_{hf}$ hrs.}
	\label{kpitiretemp}
	\vspace{-0.5cm}
\end{figure}

%
Positive objective scores in Figure \ref{kpitiretemp} show all targets are achieved regardless of the tyre temperature constraints. 
From the tuning experiments, the performance of \fastcon{} in terms of objective score and soft constraint satisfaction peaks when  $4\text{ hrs}\leq H\leq8\text{ hrs}$ with $H_{0.5}=1.2$ hrs or $\zeta=0.97$.
High queuing durations corresponding to low hot tyre durations within these horizons show that \fastcon{} controlled the tyre temperature by making smart dispatch decisions, without relying on parking actions.
This is because MCTS is able to trade off between the opportunity costs of satisfying constraint.
A prominent decrease in performance is observed with large horizons and large discount factors, as MCTS cannot accurately trade off opportunity costs as delays are not discouraged.
A high duration in hot tyres is observed with \fasthc{} as it only controls the hard constraint $ y_i < Y_{\max}$.  
The best tuned \fastcon{} spends 96\% lesser time with hot tyres than \fasthc, with 1\% improvement in terms of objective score.
%

\subsubsection{Capacity and Ratio Constraints}
\textit{i) Volume at $\text{UL}_1 > V_{\text{min}}$;
	ii) Material volume ratio at $\text{UL}_1$ is $V_{\text{M}_1}:V_{\text{M}_3} = 1:2$.}

As tuning trends are similar, results have been tabulated in Table \ref{tuningtable} to avoid duplicating graphs.
\fastcon{} was able to achieve targets with $H>1.6$ hrs and $\zeta > 0.5$.  
Results show comparable performance of best tuned \fastcon{} and \fasthc{} in terms of objective score, material moved, maintaining zero hard constraint violation and minimising soft constraint violation.
\fastcon{} achieves 5\% reduction in queuing time and consequent increase in operational time. 
Thus, \fastcon{} minimises constraint violation by effectively sequencing and scheduling the dispatch decisions, unlike \fasthc{} which enforces queuing for constraint satisfaction. 
\subsubsection{Discussion}
Table \ref{tuningtable} presents the range of \horizon{} and \halftime{} for maximising objective and ensuring constraint satisfaction, and online execution time for a single decision with maximum branching.
The execution time of \fastcon{} is determined by the number of actions to select from, efficiency of heuristics and horizon length in MCTS rollouts.
%
Note that the optimised version of FAST with no constraints in \cite{Seiler2022} takes less than 2.5s for a decision in an environment with 8 loading and 8 unloading stations, 30-40 trucks and $H\!=\!40$ mins.
Tuning results in Table \ref{tuningtable} indicate the acceptable ranges of horizon and half-times are constraint dependant and can be in conflict with other constraints.
Running arbitrarily large horizons with large discount factors lead to delayed reward accumulation, inaccurate opportunity cost estimation and noisy rewards if the environment is subject to change.
This is undesirable for MCTS rollouts as it can cause a biased exploration resulting in sub-optimal solutions.
Thus, combining multiple constraints together is non-trivial with opportunity cost formulation of constraint satisfaction.
Moreover, dependence on long time horizons for generating accurate opportunity costs makes the MCTS rollouts expensive. 
Parellelization of MCTS will be required to make algorithm scalable.
A drawback of utilising generator models to create the opportunity cost is that the trade-off between constraint satisfaction and maximising objective cannot be controlled.

\section{Conclusion}
In this article, MCTS based FAST was extended to \fastcon{} by incorporating operational constraint satisfaction.
Opportunity costs on the objective function were utilised to model constraint violations and satisfaction. 
Given an objective in terms of cumulative error, heuristic delay functions were designed to estimate opportunity costs for a variety of constraints by leveraging generator models in MCTS.
%
The success in utilising opportunity costs for constraint satisfaction was demonstrated empirically.
Performance improvements of \fastcon{} over \fasthc{} motivate the need to integrate operational constraint satisfaction into dispatch planning to improve overall production rates and prolong life of equipment. 
In future work, we aim to develop strategies to combine multiple operational constraints while maintaining scalability and test on data from real mine operations.
\begin{table}[t]
	\caption{Tuned parameters and performance of \fastcon}
	\label{tuningtable}
	\centering
	\vspace{-0.3cm}
	\begin{tabular}{rlllll}
		\textbf{Constraints} & \multicolumn{2}{c}{\textbf{Horizon} (hrs)} & \multicolumn{2}{c}{\textbf{Halftime} (hrs)} & \textbf{Comp. time}\\
		& Min. & Max.  & Min. & Max. & Max.\\
		None & 0.6 & 2.6  & 0.6 & 1.6 & 22.4s\\
		Battery & 10.5 & 14  & 3.5 & 14 & 263.4s\\
		Tyre temperature  & 4 & 8 & 1.2 & 4 & 102.3s\\
		Ratio \& capacity & 1.6 & 2.6 & 0.6 & 2.6 & 59s\\
	\end{tabular}
\vspace{-0.5cm}
\end{table}

\section*{Acknowledgement}
This work has been supported by the Australian Centre for Field Robotics and Rio Tinto Centre for Mine Automation.

\bibliographystyle{IEEEtran}
\bibliography{references_all}
\end{document}